\documentclass[journal,comsoc]{IEEEtran}

\usepackage[T1]{fontenc}
\usepackage{cite}
\usepackage{amsmath,amssymb,amsfonts}
\usepackage{algorithmic}
\usepackage{graphicx}
\usepackage{textcomp}
\usepackage{xcolor}
\usepackage{multirow}
\usepackage{booktabs}
\usepackage[colorlinks,linkcolor=blue]{hyperref}
\usepackage{fancyhdr}

\ifCLASSINFOpdf

\else

\fi

\usepackage{amsmath}

\usepackage[cmintegrals]{newtxmath}

\begin{document}

\title{Learning Dual Semantic Relations with Graph Attention for Image-Text Matching}

\author{Keyu Wen, Xiaodong Gu, and Qingrong Cheng
\thanks{This work was supported in part by National Natural Science Foundation
of China under grants 61771145 and 61371148. (Corresponding author:
Xiaodong Gu. Email: xdgu@fudan.edu.cn)
}
\thanks{The authors are with Department of Electronic Engineering,
Fudan University, Shanghai 200433, China. Email: kywen19@fudan.edu.cn.}
}

\maketitle

\thispagestyle{fancy}
\fancyhead{} 
\lhead{} 
\lfoot{} 
\cfoot{Copyright © 2020 IEEE. Personal use of this material is permitted. However, permission to use this material for any other purposes must be obtained from the IEEE by sending an email to pubs-permissions@ieee.org.} 
\rfoot{}
\begin{abstract}
Image-Text Matching is one major task in cross-modal information processing. The main challenge is to learn the unified visual and textual representations. Previous methods that perform well on this task primarily focus on not only the alignment between region features in images and the corresponding words in sentences, but also the alignment between relations of regions and relational words. However, the lack of joint learning of regional features and global features will cause the regional features to lose contact with the global context, leading to the mismatch with those non-object words which have global meanings in some sentences. 
In this work, in order to alleviate this issue, it is necessary to enhance the relations between regions and the relations between regional and global concepts to obtain a more accurate visual representation so as to be better correlated to the corresponding text.
Thus, a novel multi-level semantic relations enhancement approach named  \emph{Dual Semantic Relations Attention Network(DSRAN)} is proposed which mainly consists of two modules, separate semantic relations module and the joint semantic relations module. DSRAN performs graph attention in both modules respectively for region-level relations enhancement and regional-global relations enhancement at the same time. With these two modules, different hierarchies of semantic relations are learned simultaneously, thus promoting the image-text matching process by providing more information for the final visual representation. Quantitative experimental results have been performed on MS-COCO and Flickr30K and our method outperforms previous approaches by a large margin due to the effectiveness of the dual semantic relations learning scheme. Codes are available at \href{https://github.com/kywen1119/DSRAN}{https://github.com/kywen1119/DSRAN}.
\end{abstract}

\begin{IEEEkeywords}
cross-modal retrieval, graph attention, semantic relation, image text matching
\end{IEEEkeywords}

\IEEEpeerreviewmaketitle

\section{Introduction}

\IEEEPARstart{W}{ith} the rapid development of information technology, people's daily life is full of data of various modalities, so cross-modal information processing is increasingly important. 
For all information, visual and textual forms occupy a dominant position, thus attracting researchers to focus on cross-modal practical tasks of vision and language.
For example, the most compelling three hotspots are cross-modal retrieval \cite{cca,acmr,vse++,peng2017overview}, visual question answering \cite{vqa,bottom-up} and image captioning\cite{img-caption1,img-caption2}. In this paper, we concentrate on a subtask of cross-modal retrieval that focuses on images and texts named image-text matching. Image text matching can be studied as an independent task or as a solution to other upper level tasks. For example, Huang et al.\cite{huang2020visual} adopted image-text matching to do visual-textual reasoning for recognizing cross-media entailment (RCE). By matching specific regions and words, regions in the raw image are assigned with different labels, which benefits fine-grained visual categorization\cite{he2019fine}. Different from traditional single-modal retrieval, image-text matching \cite{lifeifei} requires the retrieval from image to text and vice versa, which is to find the most relevant text given the query image named image-based text retrieval or to find the semantically most similar image with the query text which is text-based image retrieval.

\begin{figure}[t]
\centerline{\includegraphics[width=0.9\linewidth]{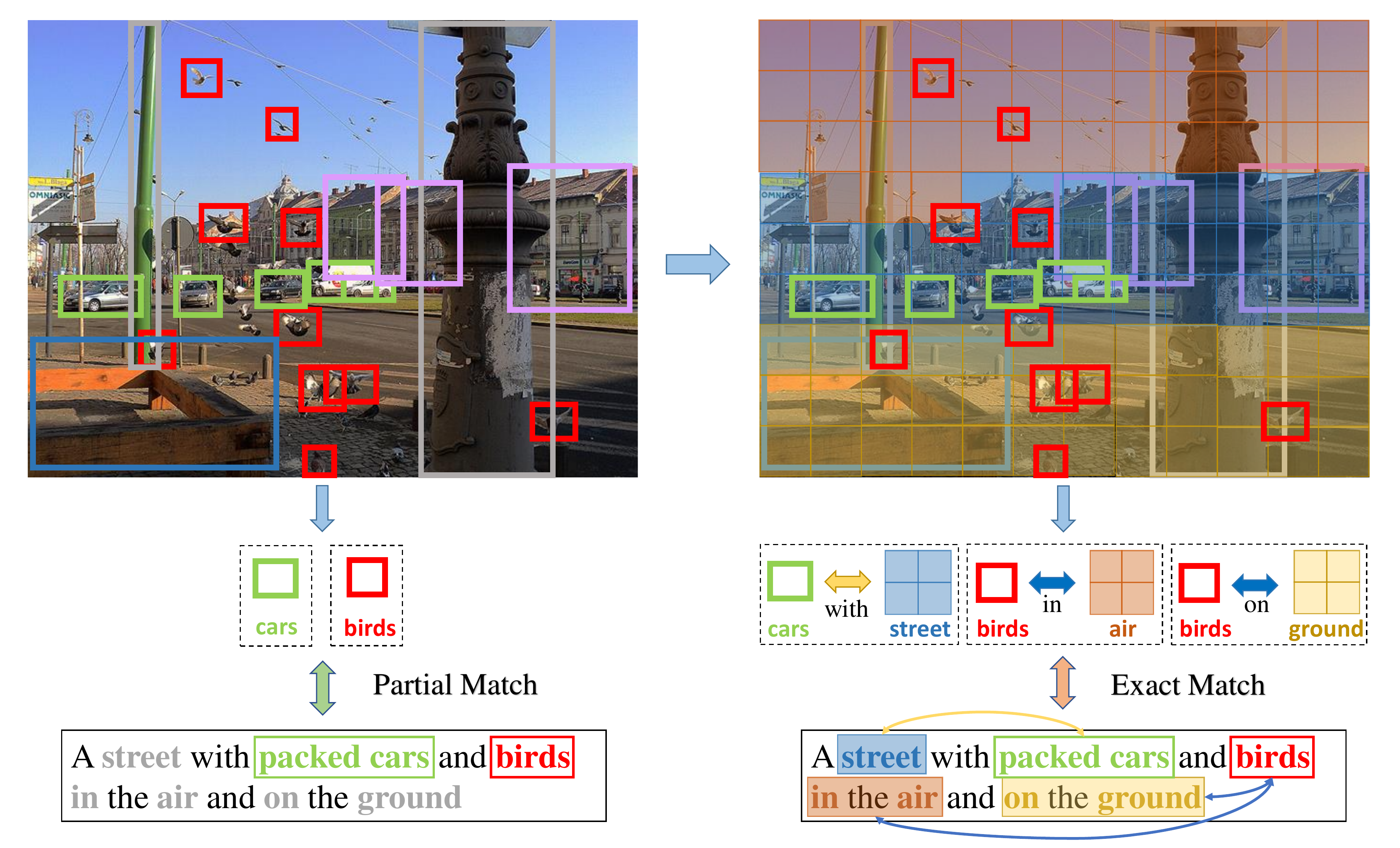}}
\caption{The proposed DSRAN learns semantic relations between regional objects as well as objects and global context. Salient objects are marked with different colored boxes. Grid shaped shadow in the right picture denotes global context such as ``air'', ``street'' and ``ground''. With only regional features, the visual representations fail to match the corresponding words and relations like ``birds \textbf{in} the air'', ``birds \textbf{on} the ground'' or ``street \textbf{with} cars''. }
\label{fig}
\vspace{-5mm}
\end{figure}

The core challenge for image-text matching is to learn a subspace where the similarities of encoded images and text representations can be directly computed, and the similarities of most correlating image-text pairs are maximized. To begin with, canonical correlation analysis (CCA)\cite{cca} stands as a backbone for this task by using a linear projection to project images and texts into the subspace. Recently with the development of deep neural networks, \cite{acmr,vse++,dcca,vse} utilize DNN to do this task. Traditionally image and text inputs are separately encoded into global visual or textual features by convolution-based networks \cite{vgg,alexnet,resnet} and RNN-based networks like LSTM\cite{lstm} or GRU\cite{gru}, after which a similarity function is used to measure the distance between two-modal representations. More recently, text representations can be obtained with pre-trained transformer-based models like BERT\cite{bert}. Pre-trained language models contain prior semantic information, which is comparable to the pre-trained CNNs in the image channel. At last, a triplet-based ranking loss function \cite{vse++} supervises the training and the best unified latent space is learned. Since these methods only take global features into account, concerning only the alignment of the whole image and the sentence, they lack a more detailed match, that is, a match between the image patches and the words in sentences.

A more refined way is to extract the local regional features using object detection methods such as Faster R-CNN\cite{faster}, which is called bottom-up attention\cite{bottom-up} for cross-modal tasks. With a pre-trained Faster-RCNN, objects in an original image can be detected. Regional features are extracted from these objects by the backbone CNNs like ResNet101\cite{resnet}. SCAN\cite{scan} firstly introduces this scheme into the image-text matching task and designs stacked cross-modal attention to align the regions in images and words in sentences. The image-text similarities are integrated from all the region-word pairs. VSRN\cite{vsrn} takes a step further to learn the relations between objects in the raw image using graph convolutional network. Since relations learning shows
an important status in image-text matching, scene-graph based methods \cite{scene-graph1,scene-graph2,scene-graph3} introduce scene graph generator to generate visual and textual scene graphs for better phrase matching. Although quite successful, these methods seem deficient at two important points: a) The alignment of regions and words seems too single for image-text matching, while the global matching process should be considered. b) They lack emphasis on relations between objects and non-object elements like the background, the surroundings, or the environment which have a strong relation to the understanding of an image when trying to match the corresponding text. As illustrated in Fig.~\ref{fig}, without learning such relations, only salient objects ``birds'' and ``cars'' are matched with the caption. However, learning the relations between objects and global context helps to assign more detailed attributes to the salient objects. For example, by adding the global-region relation ``in the air'' to the original object ``birds'', the visual representation can exactly match its textual counterpart. 

\begin{figure}[t]
\centerline{\includegraphics[width=1\linewidth]{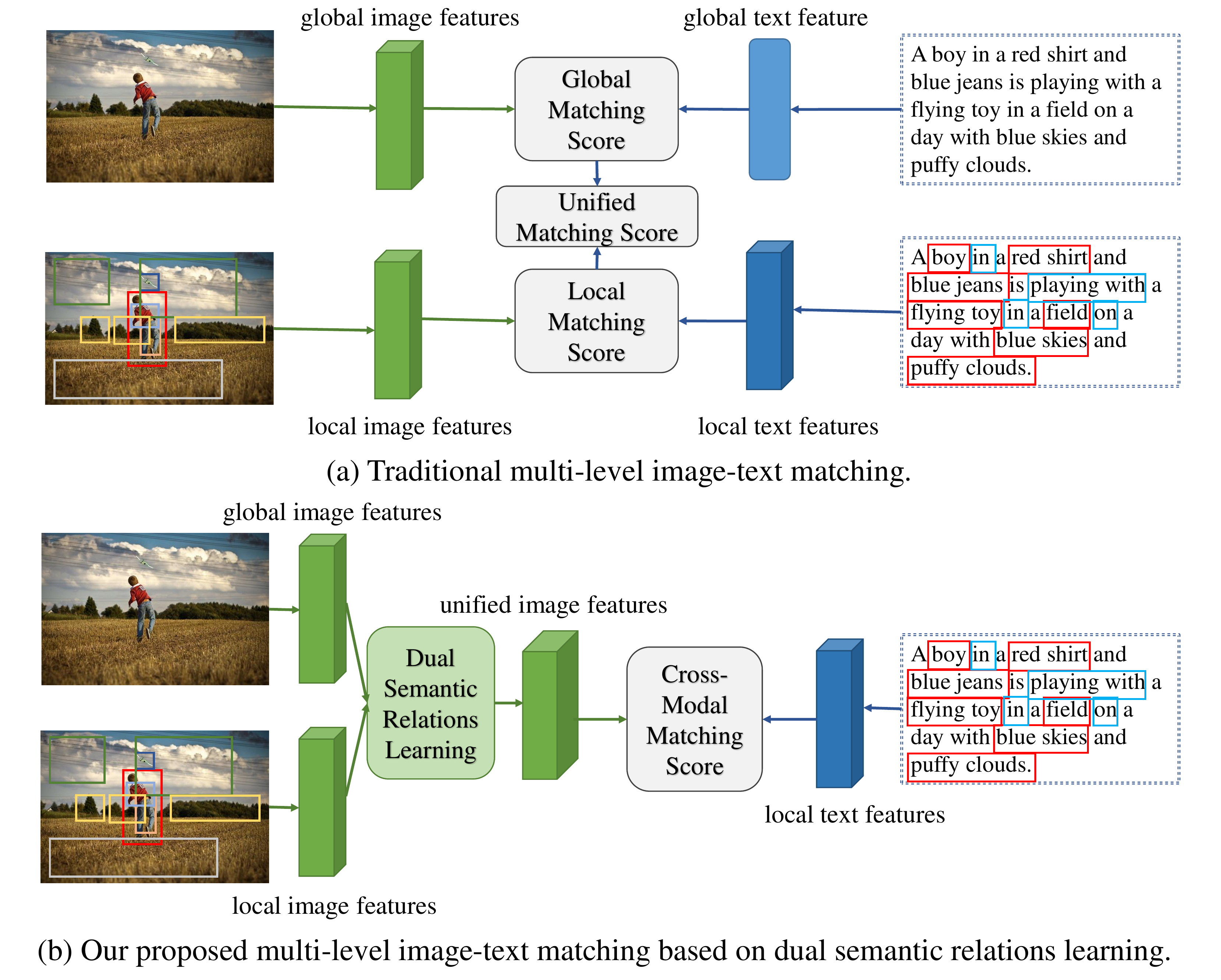}}
\caption{Different architectures of (a) traditional multi-level image-text matching methods and (b) our proposed multi-level image-text matching based on dual semantic relations learning. Objects and relations in the caption are highlighted with red and blue boxes.}\label{prob}
 \vspace{-5mm}
\end{figure}

For problem a, multi-level matching methods alleviate this by integrating both local similarities and global similarities \cite{cran,gsls,mdm,bssan,casc}, as shown in Fig.~\ref{prob} (a). They either boost the combination of image and text information by extracting both local and global features of images and captions and learning their similarities simultaneously or use a multi-label extraction scheme to get global concepts for global alignment based on regional matching methods. However for problem b, whether for global, regional, or multi-level matching methods, they all miss the attention on relations between regional objects and global concepts.

Thus, based on previous work, this paper proposes a Dual Semantic Relations Attention Network (DSRAN) to address this problem. Intuitively there are two main modules in DSRAN, the separate semantic relations module and the joint semantic relations module. The separate semantic relations module is designed for capturing regions' semantic relations. Specifically, because of the efficiency and effectiveness of GATs\cite{gat} when learning the nodes relations, the module uses two separate graph attention networks to learn pixel-wise semantic relations and regional relations at the same time. The second module, the joint semantic relations module, aims to find the semantic relations between local objects and global pixel-wise concepts. A unified graph attention network is used to achieve this. After these two principal relation-oriented modules, a gated fusion process helps to select more useful information for the final visual representation. In the end, the similarity scores of the obtained image features and text features can be calculated for further computing the value of loss function and updating the network parameters as previous works did. Details will be discussed later.

To verify our proposed model's validity, we test our model on both MSCOCO\cite{mscoco} and Flickr30K\cite{flickr} datasets. Experimental results show that our model outperforms the current state-of-the-art methods on both datasets, proving the effectiveness of our design.

Our contributions are summarized below.

(a) We propose a novel Dual Semantic Relations Attention Network(DSRAN) in order to strengthen the relations between regional objects and global concepts in the learned visual representations while considering the relations among objects themselves at the same time. 

(b) We propose a new way to learn both global and local consistency by learning a global-regional unified visual representation, instead of learning global and local similarities respectively.

(c) The proposed DSRAN outperforms previous works on the image-text matching task. Specifically, on MSCOCO our model outnumbers the current best model VSRN\cite{vsrn} by 3.0\% for image retrieval and 9.2\% for text retrieval (Recall@1 using 5K test set). Moreover, on Flickr30K, the increase is more significant, which is 8.2\% for image retrieval and 12.9\% for text retrieval (Recall@1).


\section{Related Work}\label{rw}

\subsection{Global Image-Text Matching}

Image text matching can mainly be divided into three kinds: global matching methods, regional matching methods and multi-level matching methods. Specifically, for global image-text matching methods, the goal is to embed raw images and texts into a common subspace in an end-to-end way, where similarities of the embedded visual and textual features can be directly calculated. The primary challenge lies in the respective mapping progress of images and texts.

Initially, CCA\cite{cca} utilizes a linear projection to encode cross-modal data into a common subspace where they are highly correlated. Later, researchers apply DNNs into the projection process like \cite{dcca,vse,dspe}. DCCA\cite{dcca} constructs multiple stacked layers of nonlinear transformation and learns the maximized correlations of visual and textual representations. Further in DSPE\cite{dspe}, correlation learning between cross-modal encoded features is enhanced by constructing a triplet ranking loss. However, these methods seem too plain to extract the abundant information in the images and captions. Thus, Kiros et al.\cite{vse} introduced the CNN-LSTM structure to learn a joint image-text embedding. Because of the strong ability convolutional neural networks show in image processing, a CNN pre-trained on the ImageNet dataset\cite{imagenet} takes the original image as input and outputs the encoded image features. Samely, LSTM\cite{lstm} and GRU\cite{gru} show powerful strength in natural language processing thus they are used to extract the global semantic features of the sentences. VSE++\cite{vse++} introduces the concept of hard negatives, which act as a basis for many subsequent studies. With the success of pretraining in NLP field like BERT\cite{bert} and GPT\cite{gpt}, more specific text representations can be learned as did in TOD-Net\cite{tod-net}.

Inspired by generative adversarial networks\cite{gan}, similar generative and adversarial learning schemes can be applied in image text matching task by reducing the heterogeneous gap between two modalities. CM-GANs\cite{cm-gan} and ACMR\cite{acmr} add a modality discriminator to the traditional two-way network to distinguish the modal information of the features. When unable to judge, it is considered that the heterogeneous gap between the two modalities has been eliminated. GXN\cite{gxn} generates images or captions using the learned textual or visual features, thus boosting reducing cross-modal information gap. Wen et al.\cite{cmpd} proposed a cross memory network with pair discrimination to capture the common knowledge between image and text modalities.

More special mechanisms are used in the global-wise matching. DAN\cite{dan} applies attention mechanisms for both visual and textual features enhancement. In MTFN\cite{mtfn}, Wang et al. proposed a re-ranking scheme for a more precise ranking process during testing. To achieve more comprehensive matching, MFM\cite{mfm} utilizes multi-faceted representations of image and text to characterize them more comprehensively. Thus the matching relationship between two modalities is discovered from multiple perspectives. Ji et al.\cite{san} proposed Saliency-guided Attention Network (SAN) which adopts visual saliency detection to highlight visually salient regions or objects in an image in accordance with words in sentences.

All the methods mentioned above are summarized as the global image-text matching methods as they all directly encode the whole images or texts into vectors. However, for these methods, they simply consider the alignment of the global context of images or texts while ignoring the alignment of image regions and words, which can be alleviated by the regional image-text matching methods as described below.

\subsection{Regional Image-Text Matching}\label{2-b}
A more refined way for matching images and texts is to match the salient regions in the images and the words in the sentences rather than merely matching the global semantics. Instead of traditional CNNs for image feature extraction, regional image-text matching methods utilize object detection\cite{faster} to detect the objects in the images. At the same time, the text encoder no longer outputs the global sentence vectors but word-level matrices.

Thus methods like \cite{lifeifei,scan,pfan,vsrn} take the alignment of image regions and words as the alignment of the whole image and sentence. Karpathy et al.\cite{lifeifei} first came up with a way to detect objects in images and encoded them into the subspace, where pair-wise image-text similarity is calculated by summing up similarities of all region-word pairs. SCAN\cite{scan} introduces bottom-up attention\cite{bottom-up} scheme and uses a Faster R-CNN\cite{faster} pre-trained on Visual-Genome dataset\cite{visual-genome} to encode images into region-level features while texts are encoded into word-level features. Then stacked cross-modal attention is used for similarity calculation. PFAN\cite{pfan} goes a step further by adding position information of regions into account and designs a position information integration scheme for better matching. Considering there are always words referring to relations in the sentences, VSRN\cite{vsrn} applies GCNs\cite{gcn} to do the visual reasoning and learns the relations among regions that are in accord with the text modality. Hu et al.\cite{rdan} proposed a relation-wise dual attention network (RDAN) to capture multi-level cross-modal relations with a visual-semantic relation CNN model and apply dual pathway cross-modal attention. Knowing the importance of learning relations between regions and between words for more refined matching, researchers construct scene graphs for images and sentences separately in which regions/words and relations are encoded into an exclusive graph as in \cite{scene-graph1,scene-graph2,scene-graph3}. Thus the matching process of images and texts turns into the matching of visual scene graphs and textual scene graphs which is more subtle.
In \cite{peng2019unsupervised}, the authors extended the supervised image-text matching to an unsupervised one using domain adaptation with scene graph.

\begin{figure*}[htbp]
\centerline{\includegraphics[width=1\linewidth]{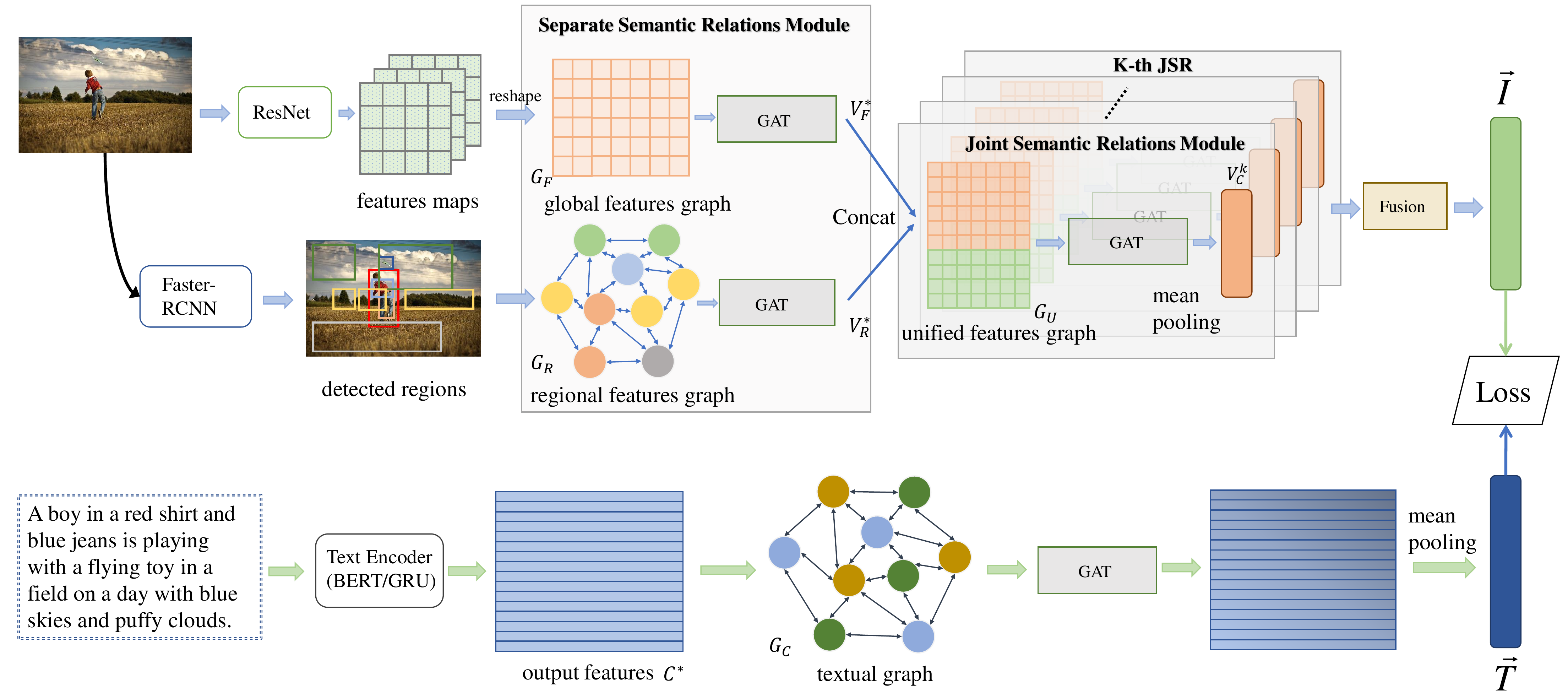}}
\caption{An overview of the proposed Dual Semantic Relations Attention Network. The model captures dual-path image features with ResNet and Faster R-CNN. Nodes of $G_F$ and $G_R$ are global regional features and local object features respectively. Two semantic relations learning modules are applied to enhance both object-level relations and unified global-regional relations. A separate or unified graph attention network is used for each of the two modules. The caption is fed to whether a pre-trained BERT-BASE encoder or a GRU encoder and we construct the textual graph with the output features where nodes of the graph are word representations. Furthermore, a textual GAT is designed for deeper textual relations learning.}\label{fig2}

\end{figure*}

\subsection{Multi-Level Image-Text Matching}
Either the global matching methods or the regional matching methods seem too single for the fine-grained matching which requires not only the alignment of global context but also the alignment of regional objects and words as well as the alignment of relations. The core concept of multi-level matching methods is to learn the correlation of both the global context and the regional/word-level concept. 

An easily thought of way is to deal with the problem in a multi-pathway as in \cite{cran,gsls,mdm,bssan}. In these works, global similarity and local similarity are calculated in separate paths and are integrated into the final similarity, as shown in Fig.~\ref{prob} (a). Specifically in MDM\cite{mdm} and GSLS\cite{gsls}, two separate paths are designed for respective global and local similarity calculations. They use either a local image CNN or a Faster RCNN to encode the local features, and use a CNN to encode global features. For texts, the original sentences are encoded into either global feature vectors or word-level matrices. Further, in BSSAN\cite{bssan} the word to regions (W2R) attention network and the object
to words (O2W) attention network are added to each path to compute attention-based matching scores. Additionally, CRAN\cite{cran} expands the network to a three-paths one by adding the relation matching path, where the relations of image patches and of words are extracted and matched.

Other methods define the global context in a more specific way. 
For example, Xu et al. \cite{casc} dealt with the global alignment of images and texts in a different way. In their model, local similarity calculation is the same as that in SCAN\cite{scan}. 
They extract semantic labels of images and sentences as the global context with a multi-label classification module for global consistency.

Although these works successfully learn multi-level consistency of images and texts for their matching, they have two apparent deficiencies: a) The design of computing global and local similarity separately limits the model to learn the relations between local objects and some global information, as illustrated in Fig.~\ref{fig}. b) Most of these works compute the final similarity by integrating global similarity, local similarity and even relation similarity, which seems too complicated for real-time applications, for it increases not only calculation complexity but also memory usage. Our work differs from previous multi-level matching methods just in the two points mentioned before. As illustrated in Fig.~\ref{prob}, our DSRAN encodes the global and local features at the same time, learns the relations among themselves or between them, and obtains the visual representation with enhanced relations. Then the model calculates the similarity of the two-modal features directly and then matches them.

\subsection{Graph Attention Network}

In this paper, we apply graph attention networks\cite{gat} to capture both object-level visual relations and global-regional visual relations. As much visual information can not always be expressed as a grid-like structure such as the visual graphs, GNNs\cite{gnn} is first introduced as a generalization of recursive neural networks which can directly process the graphs. Then GCN\cite{gcn} was proposed and further utilized in visual relations capturing as did in VSRN\cite{vsrn}. Recently an attention-based method directly using at graphs named GAT\cite{gat} overcomes disadvantages of GCN with masked self-attention mechanisms. By stacking the layers that nodes can participate in their neighborhood features, different weights can be implicitly assigned to different nodes in the neighborhood, without any type of computation-intensive matrix operation or prior understanding of the graph structure. RE-GAT\cite{regat} introduces this kind of attention based graph networks to do visual relational reasoning and promotes the cross-modal information learning in VQA.

\section{Proposed Method}
\subsection{Overview of Our DSRAN Approach}\label{overview}
In this section, we detail our proposed Dual Semantic Relations Attention Network(DSRAN). As shown in Fig.~\ref{fig2}, given an image-text pair, two separate encoding paths are designed for two modalities to get the final representations. For the image part, the raw image is firstly extracted in two levels, the global level and the object level (\ref{A}). Two modules are followed, the first of which is the separate semantic relations module aiming to learn the region-level semantic relations (\ref{B}). The second is the joint semantic relations module, which is designed for capturing relations across objects and global concepts (\ref{C}). For the text part, either a pre-trained BERT-BASE model\cite{bert} or a GRU encoder\cite{gru} extracts the representations of the words corresponding to the image features (\ref{D}). With the cross-modal representations, we can calculate the similarity scores and update the network parameters with the loss function(\ref{E}). In the end, we use a re-ranking process for more refined matching(\ref{F}).

\subsection{Two Levels of Image Features}\label{A}
Given a raw image $I$, global-level features $F$ and region-level features $R$ are extracted respectively. Generally, a ResNet152\cite{resnet} pretrained on ImageNet \cite{imagenet} whose last fully-connect layer is removed extracts the global features of the image. We use the feature map of last layer and reshape it to a set of features $F=\{f_{i},...,f_{n}\}, f_{i}\in \mathbb{R}^{D_{o}}$ where $n$ is the reshaped feature map size and $D_o$ refers to the dimension of each pixel. For the region-level part, inspired by bottom-up attention\cite{bottom-up}, the objects are firstly detected by a Faster-RCNN\cite{faster} pretrained on Visual-Genome\cite{visual-genome} dataset and then fed into a backbone Resnet101 and the output features can be represented as $R=\{r_{i},...,r_{k}\}, r_{i}\in \mathbb{R}^{D_{o}}$ where $k$ is the detected objects number. In order to embed them into the shared latent space, a fully-connect layer is carried out.
\begin{equation}
V_F = W_fF+b_f,V_R = W_rR+b_r.\label{eq1}
\end{equation}
$W_f$ and $W_r$ are the weight matrices together with the bias $b_f$ and $b_r$. Then we get the two-levels extracted features $V_F\in \mathbb{R}^{D_{e}}$ and $V_R\in \mathbb{R}^{D_{e}}$ representing visual global features and regional features where $D_e$ is the embedding dimension.

\subsection{Separate Semantic Relations Module}\label{B}
Targeting at dual-level features, we design two separate semantic relations enhancement models for learning the enhanced pixel-wise relations and object-wise relations. Specifically, we detail them in three parts, the first of which is the construction of the graph attention module.

\begin{itemize}
\item
Graph Attention Module
\end{itemize}

Given a fully-connected graph $G=(V,E)$, where $V=\{v_{i},...,v_{N}\}, v_{i}\in \mathbb{R}^{D}$ is the node features and $E$ is the edge set. Following \cite{gat}, we compute attention coefficients and normalize them with softmax function.

\begin{equation}
e_{ij} = a(W_qv_i,W_kv_j),\label{eq11}
\end{equation}

\begin{equation}
a(W_qv_i,W_kv_j) = W_qv_i(W_kv_j)^T/\sqrt{D},\label{eq33}
\end{equation}

\begin{equation}
\alpha_{ij} = Softmax(e_{ij}).\label{eq22}
\end{equation}
$W_q$ and $W_k$ are learnable parameters. In case of memory explosion, different from using the feed-forward neural network as did in the original GAT\cite{gat}, we compute the attention coefficients with multi-head dot production \cite{attention}, which is much faster and more space-efficient in practice.

\begin{equation}
MultiHead(v_i,v_j)=W_o\|_{h=1}^H(head_1,...,head_h),\label{eqh}
\end{equation}
where 

\begin{equation}
head_h = Softmax(\frac{W_q^h v_i(W_k^h v_j)^T}{\sqrt{d}})W_v^h v_j.\label{eqh2}
\end{equation}

In eq.\eqref{eqh}, $\|$ means concatenation. The projections are parameter matrices $W_q^h \in \mathbb{R}^{D\times d}$, $W_k^h \in \mathbb{R}^{D\times d}$, $W_v^h \in \mathbb{R}^{D\times d}$ and $W_o \in \mathbb{R}^{D\times D}$.

In this paper, we employ $H = 8$ parallel attention layers thus $d$ equals $D/8$. Then, with a nonlinear activation function, the final output feature can be computed. 

\begin{equation}
v_{i}^{'} = ReLU(\sum_{j\in{N_i}}MultiHead(v_i,v_j)). \label{eq44}
\end{equation}

Here $N_i$ is the neighborhood of node $i$ in the graph. We add a batch normalization into the graph attention module to accelerate training.

\begin{equation}
v_{i}^{'} =  BN(v_{i}^{'}). \label{eqbn}
\end{equation}

 In eq.\eqref{eqbn} $BN$ is the batch normalization layer. Here we finish the construction of a graph attention module.
\begin{itemize}
\item
Attention for pixel-wise relations enhancement
\end{itemize}

Obtaining the global features $V_F$, firstly we construct the global visual graph $G_F=(V_F,E_F)$. Here the edge set $E_F$ is defined as the affinity matrix by calculating the affinity edge of each pair of global features $v_F^i$ and $v_F^j$,

\begin{equation}
E_F(v_F^i,v_F^j)=(v_F^i)^T v_F^j. \label{edge1}
\end{equation}

Here more correlated image regions have edges of higher affinity scores. Thus we have the fully-connected global visual graph $G_F$.
With a graph attention module illustrated above, this process outputs global semantic-relations-enhanced features. 

\begin{equation}
V_F^*=GAT(G_F), \label{eq3}
\end{equation}
where $GAT$ means the graph attention module illustrated before. 

This progress determines how much every pixel is affected by other pixels, where semantically more corresponding pixels may have higher attention values in the image, thus promoting the pixel-wise relations learning.

\begin{itemize}
\item
Attention for object-wise relations enhancement
\end{itemize}

For more refined matching of images and texts, recent regional matching methods emphasize the importance of learning the relations of objects in raw images in alignment with text phrases. VSRN\cite{vsrn} and ML-GCN\cite{ml-gcn} illustrated the strong potential of GCNs\cite{gcn} for capturing regional relations. Different from them, this process tries to capture regional relations with a graph attention network. As seen in Fig.~\ref{fig2}, a fully-connected graph is constructed as $G_R=(V_R,E_R)$, where $V_R$ is the regional features and $E_R$ is the edge set defined as the affinity matrix defined in eq.\eqref{edge2},

\begin{equation}
E_R(v_R^i,v_R^j)=(v_R^i)^T v_R^j. \label{edge2}
\end{equation}

Graph attention networks deal with the objects graph, which contains both the object features and their relations and output semantic-relations-enhanced regional representations, as shown below.
\begin{equation}
V_R^* = GAT(G_R).\label{eq5}
\end{equation}


\subsection{Joint Semantic Relations Module}\label{C}
This part describes the kind of semantic relations that previous works lack, the object-global wise relations. As seen in Fig.~\ref{fig2}, a multi-head graph attention module is adopted with a certain purpose to bridge the relations between regional objects and global concepts. Finally, a fusion process helps to fuse the multi-head outputs and filter out more useful information.

Firstly, the enhanced global and regional features $V_F^*$ and $V_R^*$ are concatenated in the object-pixel dimension into $V_U$, $V_U=\{v_u^i,...,v_u^{n+k}\}, v_u^i\in \mathbb{R}^{D_{e}}$. And a unified features graph $G_U=(V_U,E_U)$ is obtained where $E_U$ is the edge set defined in eq.\eqref{edge4}. 
\begin{equation}
E_U(v_U^i,v_U^j)=(v_U^i)^T v_U^j. \label{edge4}
\end{equation}

Then, unified graph attention is conducted, just like in \ref{B}. Different from that in \ref{B}, here the input is the concatenated features, therefore, helping an object or a pixel learn the attention value based on all objects and pixels. With such a scheme, named joint attention, models can easily learn semantic relations between all separate elements no matter it's a regional object or a global concept.  Speciﬁcally, to stabilize the learning process of self-attention, we employ multi-head attention as did in GAT\cite{gat}. As seen in Fig.~\ref{fig2}, we feed the input $G_U$ into $K$ different GATs and the output is denoted as $V_C=\{\vec{V}_C^1,...,\vec{V}_C^K\}$. And $\vec{V}_C^k$ is defined as below.
\begin{equation}
\vec{V}_C^k = Mean(GAT_k(G_U)),\label{eq5}
\end{equation}
where $GAT_k$ refers to the graph attention module in the $k$-$th$ $JSR$(joint semantic relations module) and $Mean$ is mean-pooling.

For this module, the multi-head number $K$ is set to \{1,2,4\}. The impact of changes in $K$ will be discussed in Section \ref{V-B}. The multi-head outputs should be fused by the fusion process.

\begin{figure}[t]
\centerline{\includegraphics[width=0.9\linewidth]{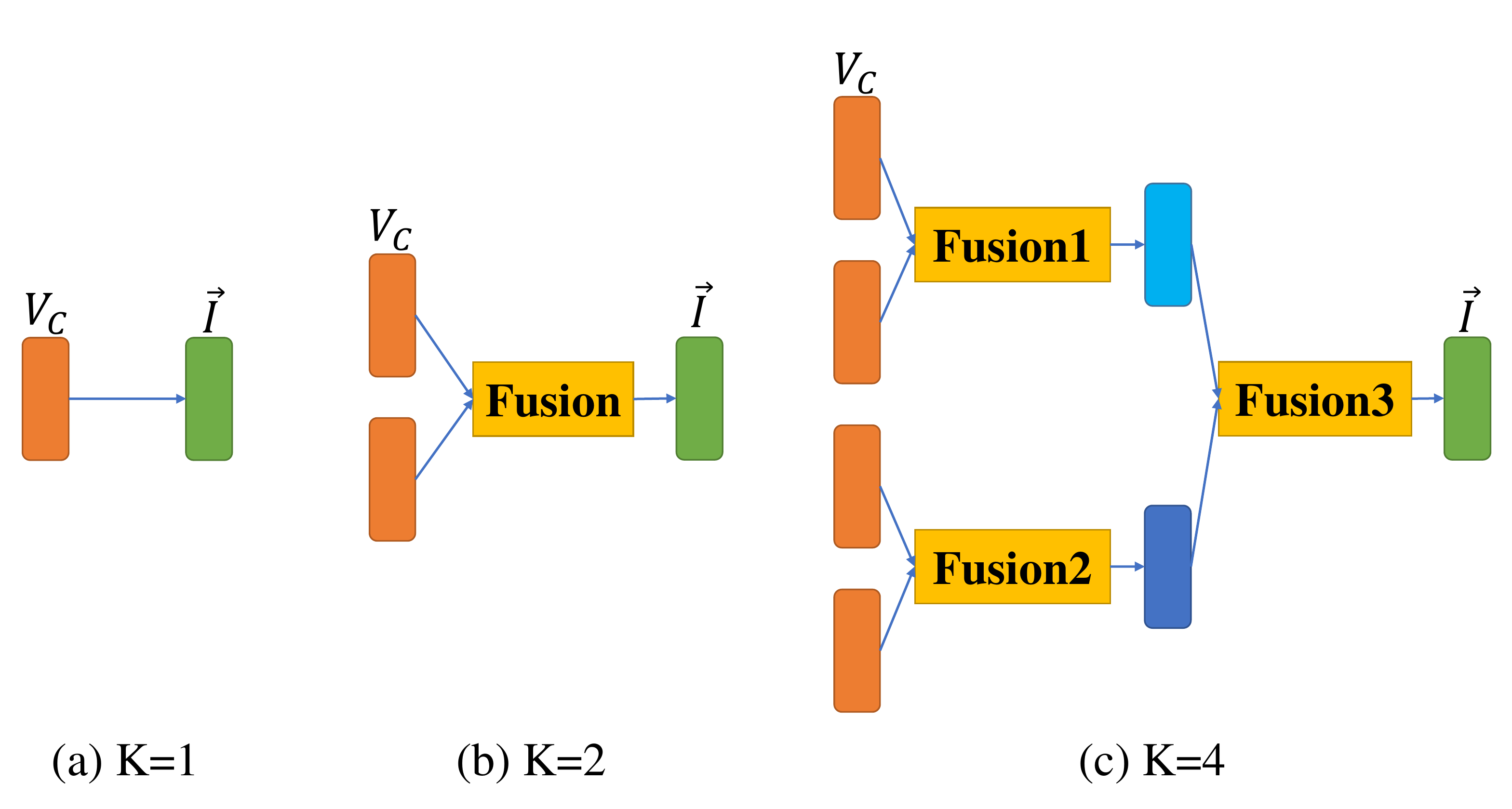}}
\caption{Illustration of different fusion process with the change of $K$.}\label{fusion}

\end{figure}

\begin{itemize}
\item
Fusion Process
\end{itemize}

With the multi-head output features $V_C$ obtained by the joint graph attention module, we fuse them with a gated fusion layer to filter more useful information and get the final image representation. 

The gated fusion layer takes two vectors $\vec{V}_C^i$ and $\vec{V}_C^j$ as input and outputs a fused representation.
\begin{equation}
\begin{aligned}
&\vec{V_1}=W_1\vec{V}_C^i,\quad \vec{V_2}=W_2\vec{V}_C^j,
\quad t=\sigma(U_1\vec{V_1}+U_2\vec{V_2}),\\
&\vec{V} = t \odot \vec{V_1}+(1-t)\odot\vec{V_2},
\quad \label{eq13}
\end{aligned}
\end{equation}
where $W$ and $U$ are the fully-connected layer parameters, $\sigma$ is the sigmoid function to project coefficients to a scale 0-1.
Considering the different values of $K$, we distinguish them in the fusion process as illustrated in Fig.~\ref{fusion}. 

$i)$ $K=1$. No fusion is needed, the final image representation is $\vec{I}=V_C$. 

$ii)$ $K=2$. Since $V_C$ has two parts $\vec{V}_C^1$ and $\vec{V}_C^2$, the final image representation comes from the output of one gated fusion process. 
\begin{equation}
\vec{I}=F(\vec{V}_C^1,\vec{V}_C^2)\label{eq14},
\end{equation}
where $F$ is the gated fusion layer as defined in eq. \eqref{eq13}.

$iii)$ $K=4$. $V_C$ has four parts $\vec{V}_C^1$, $\vec{V}_C^2$, $\vec{V}_C^3$ and $\vec{V}_C^4$. Totally we need three gated fusion layers.

\begin{equation}
\vec{I}=F_3(F_1(\vec{V}_C^1,\vec{V}_C^2),F_2(\vec{V}_C^3,\vec{V}_C^4))\label{eq15},
\end{equation}
where $F_1,F_2$ and $F_3$ mean the three gated fusion layers.

\subsection{Learning Text Representation}\label{D}

Given the original sentence $T$ corresponding to its matching image, the deep neural network embeds it into word representations. Traditionally, an RNN based network like LSTM\cite{lstm} or GRU\cite{gru} is used to process the embedded word vectors, and the output hidden states are regarded as the word representations. Recently with the development of the pre-training scheme in the NLP field, another more sophisticated substitute is to use BERT\cite{bert} as the text encoder. The self-attention based transformer structures boost the representation learning of words, for the transformer structure is better at learning semantic consistency, especially when the sentence is quite long. In this paper, we adopt either GRU\cite{gru} or BERT\cite{bert} to learn word representations. Assume the maximum word number is $m$, so the words can be illustrated as $W=\{w_i,...,w_m\}, w_i\in \mathbb{R}^{D_w}$. Then we feed them into $i)$ the BERT-BASE encoder, which has 12 layers, and we extract the outputs of the last layer as the word representations $C$, $ii)$ the GRU encoder, and the output is the representations of the words $C$, which is a matrix whose first dimension is the maximum words number and the second dimension is denoted as $D_w$. A fully-connected layer is applied to embed the features into the shared latent space where the dimension is $D_e$.

\begin{equation}
 C^* = W_cC+b.\label{eq16}  
\end{equation}

For the text part, we construct the textual graph $G_C=(C^*,E_C)$ where $C^*$ serves as the nodes and $E_C$ is defined as the edge set.

\begin{equation}
E_C(C^*_i,C^*_j)=(C^*_i)^T C^*_j. \label{edge3}
\end{equation}

We conduct the same graph attention as in \ref{B} to obtain finer text representation. As words in sentences always have close relations with others like ``a boy \textbf{in} a red shirt'', with the graph attention module, relations between nodes in the $G_C$ are strengthened to some extend. The final text representation $\vec{T}$ can be obtained as below.

\begin{equation}
 \vec{T} = Mean(GAT(G_C)),\label{eq10}  
\end{equation}
where $GAT$ is the graph attention module in \ref{B} and $Mean$ refers to mean pooling on the word level. 

\subsection{Matching Process and Loss Function}\label{E}
After obtaining the two-modal representations $\vec{I}$ and $\vec{T}$, a hinge-based triplet ranking loss\cite{vse++} is adopted to supervise the latent space learning procedure. The loss function tries to find the hardest negatives in a mini-batch, which form the triplets with the positive ones and ground truth query. The loss function is defined below.
\begin{equation}
\begin{aligned}
L=&[\alpha+S(\vec{I^{'}}, \vec{T})-S(\vec{I}, \vec{T})]_+ + \\
&[\alpha+S(\vec{I}, \vec{T}^{'})-S(\vec{I}, \vec{T})]_+. \label{eq11} 
\end{aligned}
\end{equation}

Here $S(\cdot)$ refers to similarity function which is cosine similarity in our model. $[x]_+\equiv max(x,0)$ and $\alpha$ is the margin.

\subsection{Testing Stage with Re-ranking Scheme}\label{F}
 As did in previous works for the testing stage, the model encodes the images and texts into visual and textual feature vectors. By computing the similarities with cosine similarity, we get the similarity matrix for all the testing images and texts. Inspired by the re-ranking scheme proposed by MTFN\cite{mtfn}, a re-ranking process reorganizes the similarity matrix to get a more accurate one. Retrieval results can be easily obtained by ranking the similarities between the query and its search items. In MTFN, as text-to-image re-ranking needs another text encoding path to compute single-modal text similarity, we merely use the image-to-text re-ranking in the experiments.

\begin{table*}[t]\setlength{\tabcolsep}{3pt}
\caption{Results on MS-COCO Dataset. Methods are divided into three categories, global matching methods, regional matching methods, and multi-level matching methods that unify the global and local concepts. We give out both performances on a single model or two-models ensemble. The best results are in bold, while the suboptimal values are underlined.}
    \centering
    \begin{tabular}{lccccccc|ccccccc}
        \toprule
        \multirow{3}*{\bfseries Methods} &
        \multicolumn{3}{c}{\bfseries Image-to-Text} & \multicolumn{3}{c}{\bfseries Text-to-Image}&
        \multicolumn{1}{c}{\bfseries }&
        \multicolumn{3}{c}{\bfseries Image-to-Text} & \multicolumn{3}{c}{\bfseries Text-to-Image}\\

        \cmidrule{2-15} &
        \multicolumn{7}{c}{\bfseries 1K Test Set} & \multicolumn{7}{c}{\bfseries 5K Test Set}\\

        \cmidrule{2-15} &R@1&R@5&R@10&R@1&R@5&R@10&Rsum&R@1&R@5&R@10&R@1&R@5&R@10&Rsum\\

        \midrule
        \multicolumn{8}{l}{\bfseries Global Matching Methods} \\
        \midrule
        DCCA &22.5&34.6&45.5&19.2&30.4&41.3&193.5&6.9&21.1&31.8&6.6&20.9&32.2&119.5\\
        DSPE &50.1&79.7&89.2&39.6&75.2&86.9&420.7&-&-&-&-&-&-&-\\
        MFM &58.9&86.3&92.4&47.7&81.0&90.9&457.2&-&-&-&-&-&-&-\\
        VSE++ &64.6&90.0&95.7&52.0&84.3&92.0&478.6&41.3&71.1&81.2& 30.3 &59.4&72.4&355.7 \\
        GXN & 68.5&-&97.9&56.6&-&94.5&-&42.0&-&84.7&31.7&-&74.6&-\\
        MTFN(re-rank) &74.3&94.9&97.9&60.1&89.1&95.0&511.3&48.3 &77.6 &87.3 &35.9 &66.1& 76.1&391.3 \\
        TOD-Net(BERT-Large) &75.8 &95.3 &98.4 &61.8 &89.6 &95.0 & 515.9&-&-&-&-&-&-&- \\

        \midrule \multicolumn{12}{l}{\bfseries Regional Matching Methods} \\
        \midrule
        SCAN & 70.9 &94.5 &97.8 &56.4 &87.0 &93.9&500.5& 46.4 &77.4 &87.2 &34.4 &63.7 &75.7&384.0\\
        RDAN &74.6&\textbf{96.2}&\underline{98.7}&61.6&89.2&94.7&515.0&-&-&-&-&-&-&-\\
        PFAN & 75.8 &95.9& \textbf{99.0} &61.0 &89.1 &95.1&515.9&-&-&-&-&-&-&-\\
        SGM &73.4&93.8&97.8&57.5&87.3&94.3&504.1& 50.0&79.3 &87.9& 35.3& 64.9 &76.5&393.9   \\
        \midrule
        \multicolumn{8}{l}{\bfseries Multi-Level Matching Methods} \\
        \midrule
        
        MDM &54.7&84.1&91.9&44.6&79.6&90.5&445.4&-&-&-&-&-&-&-
 \\
        GSLS &68.9&94.1&98.0&58.6&88.2&94.9&502.7&-&-&-&-&-&-&-\\
        CSAC &72.3&96.0&\textbf{99.0}&58.9&\underline{89.8}&\textbf{96.0}&512.0&47.2&78.3&87.4&34.7&64.8&76.8&389.2\\
        \textbf{DSRAN(GRU)}&76.3&94.9&98.4&\underline{62.4}&89.7&95.2&516.9&51.9&81.6&89.8&\underline{39.5}&\underline{70.6}&\underline{81.0}&414.4\\
        \textbf{DSRAN(GRU)(re-rank)}&\textbf{79.0}&\textbf{96.2}&98.5&\underline{62.4}&89.7&95.2&\underline{521.0}&\underline{55.4}&\textbf{84.4}&\textbf{91.0}&\underline{39.5}&\underline{70.6}&\underline{81.0}&\underline{421.9}\\
        \textbf{DSRAN(BERT)}&77.1&95.3&98.1&\textbf{62.9}&\textbf{89.9}&\underline{95.3}&518.6&53.7&82.1&89.9& \textbf{40.3} &\textbf{70.9}&\textbf{81.3}&418.2\\
        \textbf{DSRAN(BERT)(re-rank)} &\underline{78.8}&\underline{96.1}&98.5&\textbf{62.9}&\textbf{89.9}&\underline{95.3}&\textbf{521.5}&\textbf{56.3}&\underline{84.2}&\underline{90.7}& \textbf{40.3} &\textbf{70.9}&\textbf{81.3}&\textbf{423.7}\\
        \midrule
        \multicolumn{12}{l}{\bfseries Two-Models Ensemble} \\
        \midrule
        SCAN &72.7 &94.8 &98.4 &58.8 &88.4 &94.8&507.9&50.4 &82.2 &90.0 &38.6 &69.3 &80.4&410.9\\
        PFAN &76.5 &\underline{96.3} &\textbf{99.0} &61.6 &89.6 &95.2&518.2&-&-&-&-&-&-&-\\
        VSRN &76.2 &94.8 &98.2 &62.8 &89.7 &95.1&516.8&53.0 &81.1 &89.4 &40.5 &70.6 &81.1&415.7\\
        TOD-Net(BERT-Large) &78.1 &96.0 &98.6 &63.6 &\underline{90.6} &\textbf{95.8} &522.7&-&-&-&-&-&-&-\\
        
        \textbf{DSRAN(GRU)}&78.0&95.6&98.5&\underline{64.2}&{90.4}&\textbf{95.8}&522.5&54.4&{83.5}&91.3&\underline{41.5}&\underline{71.9}&\underline{82.1}&424.7\\
        \textbf{DSRAN(GRU)(re-rank)}&\underline{80.4}&\textbf{96.7}&\underline{98.7}&\underline{64.2}&{90.4}&\textbf{95.8}&\underline{526.2}&\underline{57.6}&\textbf{85.6}&\underline{91.9}&\underline{41.5}&\underline{71.9}&\underline{82.1}&\underline{430.6}\\
        
        \textbf{DSRAN(BERT)} &78.3&95.7&98.4&\textbf{64.5}&\textbf{90.8}&\textbf{95.8}&523.5&55.3&83.5&90.9&\textbf{41.7}&\textbf{72.7}&\textbf{82.8}&426.9\\
        
        \textbf{DSRAN(BERT)(re-rank)}&\textbf{80.6}&\textbf{96.7}&\underline{98.7}&\textbf{64.5}&\textbf{90.8}&\textbf{95.8}&\textbf{527.1}&\textbf{57.9}&\underline{85.3}&\textbf{92.0}&\textbf{41.7}&\textbf{72.7}&\textbf{82.8}&\textbf{432.4}\\
        \midrule

    \end{tabular}
    \centering

    \label{tab:coco}
\end{table*}

\begin{table}[t]\setlength{\tabcolsep}{2.2387pt}
 \caption{Results on Flickr30K. The configurations are the same as those of MSCOCO. TOD-Net is no longer shown here because no experiments on this dataset can be found in their paper.}
    \centering
    \begin{tabular}{lccccccccll}
        \toprule
        \multirow{2}*{\bfseries Methods} &
        \multicolumn{3}{c}{\bfseries Image-To-Text} & \multicolumn{3}{c}{\bfseries Text-To-Image}\\ 
        \cmidrule{2-8} &R@1&R@5&R@10&R@1&R@5&R@10&Rsum \\
        \midrule
        \multicolumn{8}{l}{\bfseries Global Matching Methods} \\
        \midrule
        DCCA &27.9&56.9&68.2&26.8&52.9&66.9&299.6\\
        DSPE &40.3&68.9&79.9&29.7&60.1&72.1&351.0\\
        MFM &50.2&78.1&86.7&38.2&70.1&80.2&403.5\\
        VSE++ &52.9 &80.5 &87.2& 39.6 &70.1 &79.5&409.8\\
        GXN &56.8&-&89.6&41.5&-&80.1&-\\
        MTFN &65.3 &88.3 &93.3 &52.0 &80.1 &86.1&465.1 \\
        \midrule
        \multicolumn{8}{l}{\bfseries Regional Matching Methods} \\
        \midrule
        SCAN &67.9 &89.0 &94.4 &43.9 &74.2 &82.8 &452.2\\
        RDAN &68.1&91.0&95.9&54.1&80.9&87.2&477.2\\
        PFAN &67.6 &90.0 &93.8 &45.7 &74.7 &83.6&455.4\\
        SGM &71.8&91.7&95.5&53.5&79.6&86.5&478.6 \\

        \midrule
        \multicolumn{8}{l}{\bfseries Multi-Level Matching Methods} \\
        \midrule
        MDM &44.9&75.4&84.4&34.4&67.0&77.7&384.0\\
        GSLS &68.2&89.1&94.5&43.4&73.5&82.5&451.2\\
        CASC &68.5&90.6&95.9&50.2&78.3&86.3&469.8\\
        
        \textbf{DSRAN(GRU)}&72.6&93.6&96.3&\underline{56.3}&\underline{84.0}&\underline{89.8}&492.6\\
        \textbf{DSRAN(GRU)(re-rank)}&\underline{75.7}&\underline{94.7}&\underline{96.8}&\underline{56.3}&\underline{84.0}&\underline{89.8}&497.3\\
        
        \textbf{DSRAN(BERT)} &75.3&94.4&\textbf{97.6}&\textbf{57.3}&\textbf{84.8}&\textbf{90.9}&\underline{500.3} \\
        \textbf{DSRAN(BERT)(re-rank)} &\textbf{78.6}&\textbf{95.6}&\textbf{97.6}&\textbf{57.3}&\textbf{84.8}&\textbf{90.9}&\textbf{504.8} \\
        \midrule
        \multicolumn{8}{l}{\bfseries Two-Models Ensemble} \\
        \midrule
        SCAN &67.4 &90.3 &95.8 &48.6 &77.7 &85.2 &465.0\\
        PFAN &70.0 &91.8 &95.0 &50.4 &78.7 &86.1&472.0 \\
        VSRN &71.3 &90.6 &96.0 &54.7 &81.8 &88.2&482.6 \\
        
        \textbf{DSRAN(GRU)}&74.9&94.5&97.0&\underline{58.6}&\underline{85.8}&\underline{91.3}&502.1\\
        \textbf{DSRAN(GRU)(re-rank)}&\underline{79.6}&\textbf{95.6}&97.5&\underline{58.6}&\underline{85.8}&\underline{91.3}&\underline{508.4}\\
        
        \textbf{DSRAN(BERT)} &77.8&95.1&\underline{97.6}&\textbf{59.2}&\textbf{86.0}&\textbf{91.9}&507.6 \\
        \textbf{DSRAN(BERT)(re-rank)}&\textbf{80.5}&\underline{95.5}&\textbf{97.9}&\textbf{59.2}&\textbf{86.0}&\textbf{91.9}&\textbf{511.0}\\
        \midrule

    \end{tabular}
   \label{tab:f30k}
\end{table}

\section{Experiments}
To evaluate our DSRAN on the image-text matching task, we perform several experiments on both image retrieval and text retrieval. Table \ref{tab:coco} and Table \ref{tab:f30k} are the compare results with state-of-the-art methods. 
\subsection{Datasets and Evaluation Metrics}
We apply the two publicly available datasets, Microsoft COCO\cite{mscoco} and Flickr30K\cite{flickr}. In Flickr30K, there are 31,783 images with five captions each. Following \cite{vse++}, the images are split into 29,000, 1000 and 1000 for training, validation and testing. As for the MSCOCO dataset, there are a total of 123,287 images, and every image has five description captions. As did in \cite{vse++,scan,vsrn}, the splits contain 113,287 images for training, 5000 for validation and 5000 for testing. Specifically for MSCOCO, the final results are obtained either by averaging over five folds of 1k test images (referred to as 1K test set) or by directly testing the whole 5k images (referred to as 5K test set).
For both image retrieval and text retrieval tasks, we record the results by calculating the recall at K (R@K) metrics defined as the proportion of the queries whose correct retrieved results are among the top-K ranking results. Specifically, we use R@1, R@5, and R@10 together with Rsum defined as below.
\begin{equation}
\begin{aligned}
Rsum =&\underbrace{R@1 + R@5 + R@10}_{\text{image retrieval}}+\\&\underbrace{R@1 + R@5 + R@10}_{\text{text retrieval}}.
\end{aligned}
\end{equation}

\subsection{Implementation Details}
This section gives more detailed model settings and training settings for our DSRAN in the experiments. For global-wise feature maps extraction, the raw image is firstly randomly cropped and resized to $224\times224$. Moreover, we utilize the feature map of the fourth layer of Resnet152\cite{resnet} with the feature map number $n$ set to $7\times7=49$. For region-wise object features extraction, we simply use the regional features given by \cite{vlp}, and the number of regions $k$ is 100. Both kinds of features share the same dimension $D_o$, which is 2048. As for texts, we use either a pre-trained BERT-BASE\cite{bert} model, and the embedding dimension $D_w$ is 768 or a GRU encoder\cite{gru} with a word embedding size of 300. The BERT-BASE encoder is finetuned while parameters of visual encoders ResNet152 and Faster-RCNN are fixed. The embedded latent space dimension $D_e$ is set to 1024. The multi-head number $K$ in the joint semantic relations module in BERT-based models (which means models with a BERT-BASE encoder) is set to 2 or 4 for Flickr30K and MSCOCO, respectively. For GRU-based models (which means models with a GRU encoder) $K$ is set to 2 for both datasets.

For the model with BERT, experiments are performed on at least two NVIDIA 1080Ti GPU with the batch size setting to 320 for MSCOCO and 128 for Flickr30K. We train the model with an Adam optimizer\cite{adam} with a warmup rate of 0.1 for 12 and 18 epochs for Flickr30K and MSCOCO, respectively. The learning rate is set to 2$e$-5 at first and declines by ten times every 6 or 9 epochs, half of the total training epochs. While for the GRU-based model, experiments are performed on one NVIDIA 1080Ti GPU with the batch size setting to 300 for MSCOCO and 128 for Flickr30K. We train the model with an Adam optimizer for 16 and 18 epochs for Flickr30K and MSCOCO, respectively. The learning rate is set to 2$e$-4 at first and declines by ten times every 8 or 9 epochs. 

\subsection{Comparative Experiments with State-of-the-art Methods}
We compare our DSRAN model with current state-of-the-art methods. As discussed in Section \ref{rw}, they are divided into three kinds, $i)$ global matching methods DCCA\cite{dcca}, DSPE\cite{dspe}, MFM\cite{mfm}, GXN\cite{gxn}, VSE++\cite{vse++}, MTFN\cite{mtfn} and TOD-Net\cite{tod-net}, $ii)$ regional matching methods SCAN\cite{scan}, RDAN\cite{rdan}, PFAN\cite{pfan}, VSRN\cite{vsrn} and scene-graph based method SGM\cite{scene-graph1}, $iii)$ multi-level matching methods GSLS\cite{gsls}, MDM\cite{mdm}, CASC\cite{casc}. Our DSRAN belongs to the multi-level matching methods. We record results from models with BERT or GRU. It should be noticed that TOD-Net uses the 24-layer BERT-Large model rather than our 12-layer BERT-BASE model. We record results whether with a re-ranking process as mentioned in Section \ref{F} or not. Results from a single model or two-models ensemble are both recorded here. When conducting the ensemble scheme, the similarity scores from two already trained models are averaged for the final ranking process.

\begin{itemize}
\item{Results on MSCOCO}
\end{itemize}

As shown in Table \ref{tab:coco}, the highest performance of each metric is made bold, and the suboptimal values are underlined. Our DSRAN outperforms other methods, whether using an ensemble or not, except for the $R@10$ in the 1K test set, which may be due to noise. It is noticed that even without the re-ranking scheme, our model with either BERT or GRU still gains the best performance in most of the metrics. For the 1K test set, our model exceeds the current best TOD-Net\cite{tod-net} with a BERT-Large text encoder against our BERT-BASE encoder by 3.0 and 1.0 on text retrieval and image retrieval respectively at $R@1$ (single model). From the table, performance gains of $R@5$ and $R@10$ are not as significant as that of $R@1$. This may be due to the existence of more interference sources for a given query in such a large target set. For the 5K test set, similarly, our model outnumbers the state-of-the-art VSRN\cite{vsrn} by 4.9 and 16.7 considering the $R@1(I2T)$ and $Rsum$ metric. The above outperforming proves our dual semantic relations learning scheme's effectiveness, focusing on the unified global-region visual representations learning. By applying BERT encoder, we gain a little increase in some metrics, while for others, the trend seems to be the opposite ($R@1$ and $R@5$ for single model). 

\begin{itemize}
\item{Results on Flickr30K}
\end{itemize}

Performances on Flickr30K are shown in Table \ref{tab:f30k}. Our proposed DSRAN outperforms other state-of-the-art methods by a large margin. Whether with or without the re-rank scheme, our model achieves the best results in all the metrics. Compared to the previous best model VSRN\cite{vsrn}, we increase 9.2 on text retrieval and 4.5 on image retrieval ($R@1$), with a great improvement on the Rsum metric (28.4). In case of the comparison with the current best model of the three kinds of methods, $i)$ for MTFN, our model outperforms it by 13.3 and 5.3 considering $R@1$, $ii)$ for SGM, the performance gain in $R@1$ is 6.8 and 3.8 respectively for text retrieval and image retrieval, $iii)$ for CASC, our $R@1$ outnumbers theirs by 10.1 and 7.1. The I2T performance gain seems more obvious than that of T2I performance gain, which is the result of conducting only I2T re-ranking, as stated in \ref{F}. At the same time, the performances increase seems more salient when using BERT encoder for texts, which may be due to the powerful capabilities in processing longer sentences of BERT.

\begin{itemize}
\item{Results Analysis}
\end{itemize}

From the tables, we get some more detailed findings.

a) Our model is superior to all the global matching methods in all the metrics. Since our model takes both global and local regional features into account, global matching methods that only focus on the global concept alignment seem insufficient for refined image-text matching. Further, the performance gain against regional matching methods, which only concentrate on region-word alignment, proves that the combination of global and regional alignment benefits the matching as did in our method.

b) The proposed DSRAN also outperforms previous multi-level matching methods. Although MDM, GSLS, and CASC integrate both local region-word matching and global semantics matching, they lack the process of learning relations between regions and the relations between regional objects and global concepts. 

c) Compared with VSRN, who limitedly considers learning relations between regional objects in the images, our proposed global-region relations learning boosts this task. There are always relations between salient objects and hard-to-catch global concepts in the captions. Thus our dual semantic relations learning is significant for accurate matching.

\begin{figure*}[t]
\centerline{\includegraphics[width=1\linewidth]{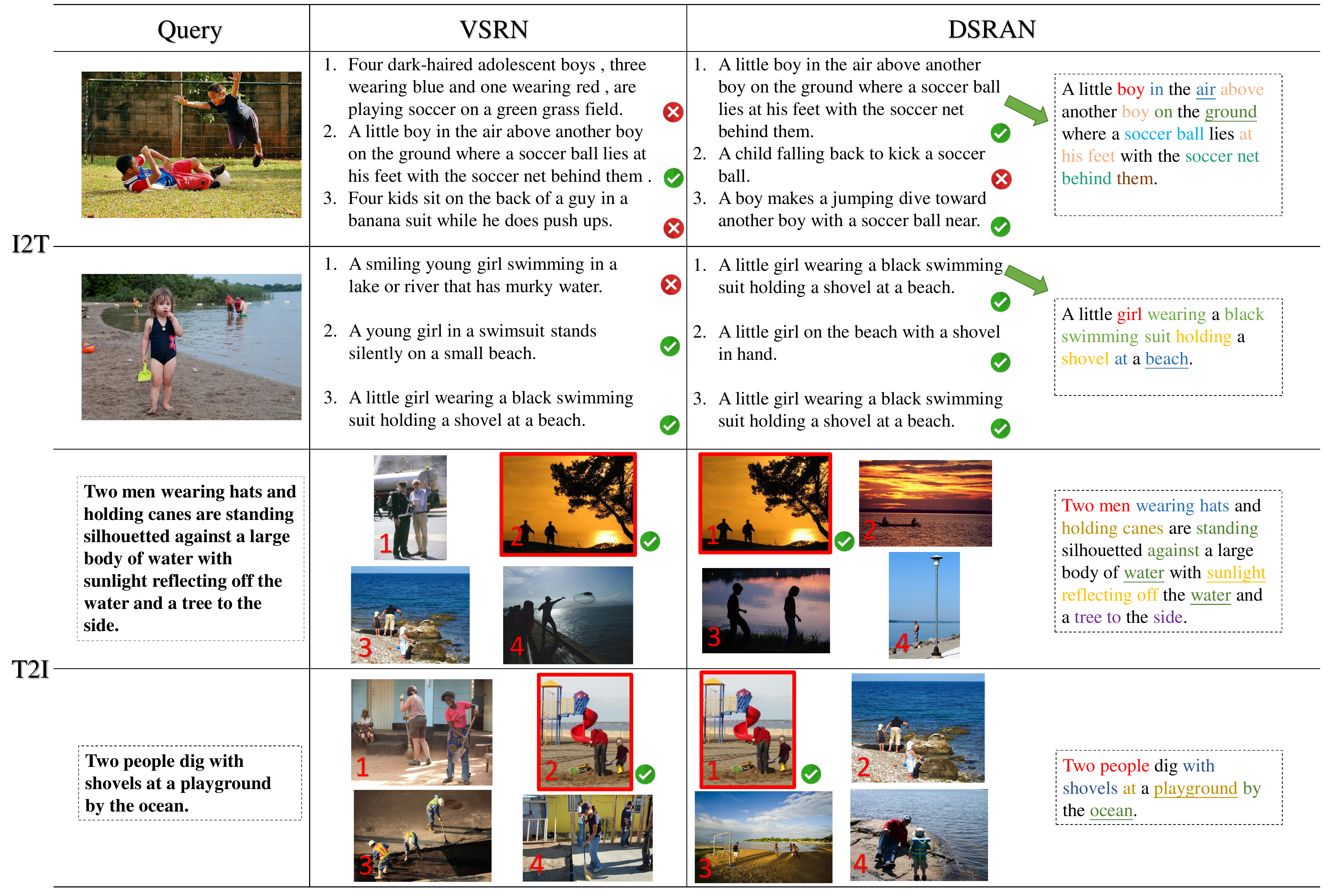}}
\caption{The retrieval results of our DSRAN and VSRN, which only considers regional relations. Experiments are conducted using the Flickr30K dataset and on the models with the highest $Rsum$. Two instances are shown for I2T retrieval and T2I retrieval each. We give the top-three ranking texts for every image query and top-four ranking images for every text query. In the figure's right column, we highlight the ground-truth caption by marking objects, global concepts, and relations with several colors. Global concepts are underlined.}
\label{rr}
\end{figure*}

\subsection{Visualization of Retrieval Results}

To validate our dual semantic relations learning scheme's practical matching results, we list the top-3 ranking items of the image query and the top-4 ranking items of the text query. We use Flickr30K dataset and the model with the highest $Rsum$ without re-ranking to do this experiment (that is DSRAN(BERT) in Table \ref{tab:f30k}). Specifically, in Fig.~\ref{rr}, we list results from two models, $i)$ VSRN\cite{vsrn} which only considers the regional relations, $ii)$ our DSRAN which integrates both regional relations and region-global relations. Here I2T means text retrieval with the image query, and for T2I, vice versa. On the illustration's far-right, we highlight the sentences that VSRN retrieves wrong, but DSRAN retrieves right. Specially, we mark the objects and global contexts together with their relations with different colors.
Moreover, the relation shares the same color with the object or global concept following it. We additionally underline the global concepts. For example, for the sentence on the second row of the table ``A little girl wearing a black swimming suit holding a shovel at a beach'',  ``girl'', ``black swimming suit'', ``shovel'' are the salient objects and ``beach'' is a global concept while ``wearing'', ``holding'', ``at'' are the relations between them.

From Fig.~\ref{rr}, it is obvious that a decisive factor that VSRN fails to match the image-text pair is that their model lacks the learning of region-global relations. As seen in the second sample, although the first retrieval caption for the query image still contains a key object with relation to the image region which is ``smiling young girl''. However, except for the salient object, global concepts and their relations with the object are equally significant. For the image, the global concept is ``beach'' and the relation is ``at''. Simply ignoring this kind of relations may lead to image-text mismatch. For example, in line one, the mismatch of VSRN comes from ``boy in the air'' and in row three, the mismatch comes from ``sunlight reflecting the water'', ``men standing against water''. And in row four, the mismatch still comes from ``at a playground'' and ``by the ocean''.

\section{Ablation Study and Analysis}
In this section, firstly in \ref{V-A} we do several ablation studies considering the dual semantic relations enhancement schemes used in our model, $i)$separate semantic relations module(SSR), $ii)$joint semantic relations module(JSR) and the integration of dual-paths including the regional path and the global path. Then in \ref{V-B}, we analyze the graph attention modules used in both SSR and JSR. At last in \ref{V-C}, we analyze the training process of our model and the effectiveness of batch normalization.

\begin{table*}[t]\setlength{\tabcolsep}{7pt}
\caption{Ablation studies on different model settings. ``Regional'' and ``Global'' refer to using whether a single-path model or both. ``SSR'' refers to the separate semantic relations module while ``JSR'' is the joint semantic relations module. ``BERT'' and ``GRU'' determine the text encoder. The top-8 rows are models with BERT encoder while the rest are those with GRU encoder.} The highest value of each metric is made bold. We run this ablation study on the Flickr30K dataset. We record results from single models without the re-ranking process.
    \centering
    \begin{tabular}{cccccccccccccc}
        \toprule
        \multicolumn{1}{c|}{\bfseries Number} &
        \multicolumn{6}{c|}{\bfseries Model Settings} &
        \multicolumn{3}{c}{\bfseries Image-to-Text} & \multicolumn{3}{c}{\bfseries Text-to-Image}\\ 
        \midrule 
        &{\bfseries Regional}&{\bfseries Global}&{\bfseries SRR}&{\bfseries JRR}&{\bfseries BERT}&{\bfseries GRU}&R@1&R@5&R@10&R@1&R@5&R@10&Rsum\\
        \midrule

        1&  &\checkmark& & &\checkmark &&48.1&78.1&87.2&36.4&68.1&78.6&396.4\\
        2&  &\checkmark&\checkmark & &\checkmark&&52.9&79.8&88.5&38.2&69.7&79.9&409.0\\
        3&\checkmark& & & &\checkmark&&58.4&84.8&91.9&44.5&75.3&84.2&439.1   \\
        4&\checkmark& &\checkmark& &\checkmark&&73.2&92.3&96.4&56.3&83.6&89.8&491.6 \\
        5&\checkmark &\checkmark& & &\checkmark&&68.0&91.7&96.0&51.6&81.2&88.4&476.9  \\
        6&\checkmark&\checkmark&\checkmark& &\checkmark&&72.8&93.4&97.1&57.1&84.5&90.8&495.6\\
        7&\checkmark&\checkmark& &\checkmark&\checkmark&&73.4&92.9&96.9&57.2&84.7&90.8&495.9\\
        8&\checkmark&\checkmark&\checkmark&\checkmark&\checkmark &&\textbf{75.3}&\textbf{94.4}&\textbf{97.6}&\textbf{57.3}&\textbf{84.8}&\textbf{90.9}&\textbf{500.3}\\
        
        \midrule
        
        9&  &\checkmark& & & &\checkmark& 43.3&73.9&83.4&33.7&63.9&74.7&372.9\\
        10&  &\checkmark&\checkmark & &&\checkmark&52.6&78.5&85.4&39.2&69.4&79.1&404.3\\
        11&\checkmark& & & &&\checkmark& 61.2&87.1&93.2&45.0&74.7&83.5&444.7
\\
        12&\checkmark& &\checkmark& &&\checkmark& 67.7&89.4&94.8&51.4&79.8&87.9&470.9\\
        13&\checkmark &\checkmark& & &&\checkmark&66.4&89.4&95.1&48.9&78.1&86.3&464.2\\
        14&\checkmark&\checkmark&\checkmark& &&\checkmark& 69.3&90.8&96.3&54.5&81.7&89.2&481.8\\
        15&\checkmark&\checkmark& &\checkmark&&\checkmark&70.1&90.7&95.5&53.2&81.8&88.9&480.2\\
        16&\checkmark&\checkmark&\checkmark&\checkmark& &\checkmark&\textbf{72.6}&\textbf{93.6}&\textbf{96.3}&\textbf{56.3}&\textbf{84.0}&\textbf{89.8}&\textbf{492.6}\\
        \midrule

    \end{tabular}
    \label{tab:sr-ablation}
\end{table*}

\subsection{Ablation Studies on DSRAN}\label{V-A}

For the visual part of our DSRAN, there are two paths, one for global features extraction and another for local regional features extraction. Thus, for ablation studies, we conduct experiments using a) only global path, b) only regional path, and c) both two paths. There are two main semantic relations modules in our DSRAN, the separate semantic relations module (referred to as SSR) and the joint semantic relations module (referred to as JSR). We perform ablation experiments on Flickr30K\cite{flickr} test set with or without the modules. Thus, there are total 8 experiments for BERT-based models and another 8 for GRU-based models as listed in Table \ref{tab:sr-ablation}. It should be noticed that no re-ranking is used in the experiments. In the table, ``Global'' refers to using the global path, and ``Regional'' refers to using the regional path. ``SSR'' and ``JSR'' refer to whether using those two semantic relations modules. ``BERT'' and ``GRU'' refer to the text encoder use in the model. For example, in line 6, ``Regional'', ``Global'', ``SSR'' and ``BERT'' are chosen, which means we use both the global and regional paths with only the separate semantic relations module and fuse them with the gated fusion layer based on BERT encoder. For the first four lines numbered 1-4 and 9-12, these are experiments using only one path. Thus no JSR is added for these four. For the last four lines numbered 5-8 and 13-16, these are experiments using both paths with or without these two semantic relations modules. It is noticed that the configurations of numbers 8 and 16 are the same as our final models for comparison. 

As shown in Table \ref{tab:sr-ablation}, for lines 1-4 and 9-12, which simply use a single path for image representation learning, single regional object-word alignment is superior to single global semantics alignment. Moreover, the use of SSR boosts the performances in a significant way proving the effectiveness of our proposed SSR aiming at learning whether pixel-wise global relations or object-wise regional relations. When using both paths in line 5-8 and 13-16, not only can we see that with our proposed dual semantic relations learning scheme, images and texts are better matched in a fine-grained way, considering both regional relations and region-global relations, but also this kind of global-regional integration scheme is superior to single-path visual encoding while computing only one similarity in the end. Comparing models with different text encoders (BERT or GRU), although GRU-based models perform lower than BERT-based models, the performance gains from the proposed modules seem more significant for the former.

\subsection{Analysis on Graph Attention Module}\label{V-B}

In both semantic relations modules, we make a little modification to traditional GATs\cite{gat} to suit our model and apply them to enhance relations learning. The model successfully learns the relations-enhanced features by constructing two separate fully-connected graphs for global features and regional features. Here we analyze GATs used in SSR and JSR, respectively.

\begin{figure*}[htbp]
\centerline{\includegraphics[width=1\linewidth]{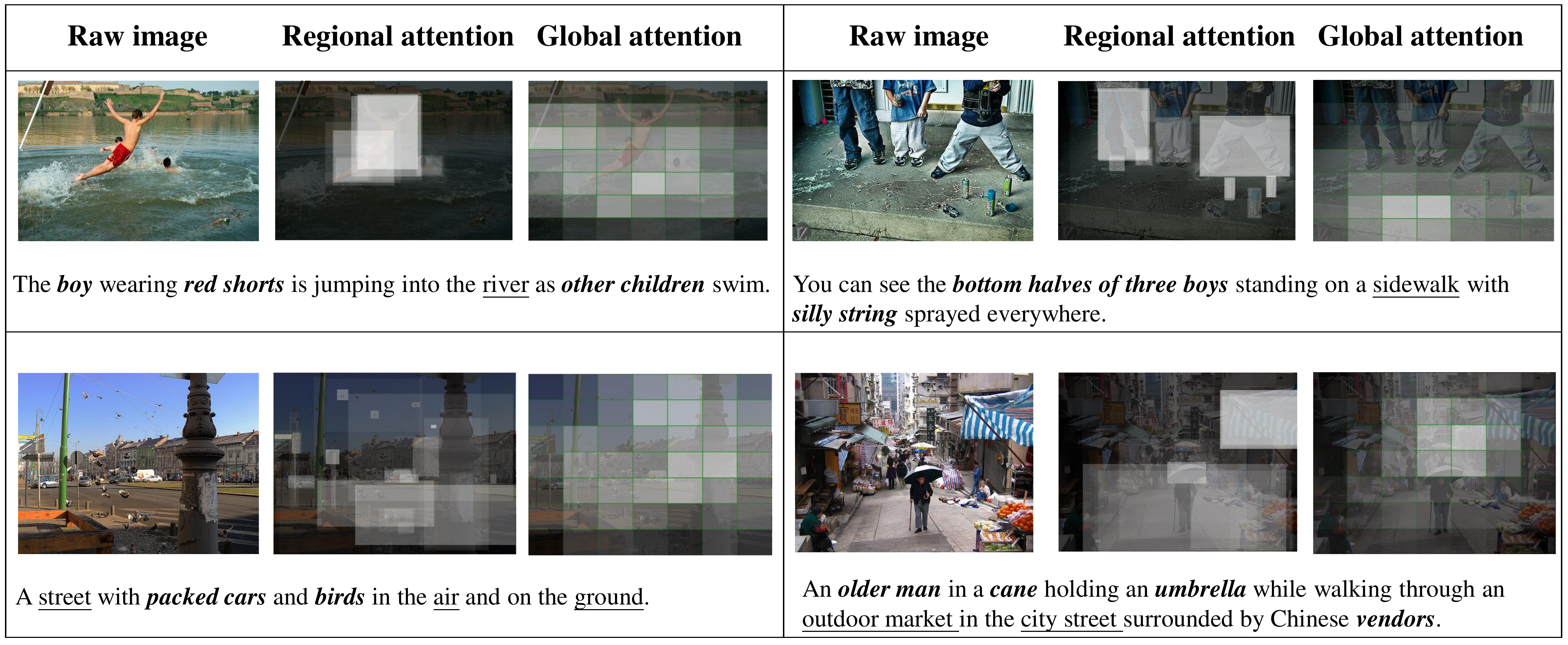}}
\caption{Visualization of regional and global attention achieved by GATs using Flickr30K dataset. The left column is the raw image, and the others are regional and global attention pictures. We have the salient objects in the texts in bold, and italics and underline global context like ``river'', ``ground'' and ``street''. In the attention pictures, more critical areas are highlighted brighter. We show the results of pictures with a brief background on the first row while pictures on the second row have more complex backgrounds.}
\label{fig5}
\end{figure*}

\begin{figure}[t]
\centerline{\includegraphics[width=1.0\linewidth]{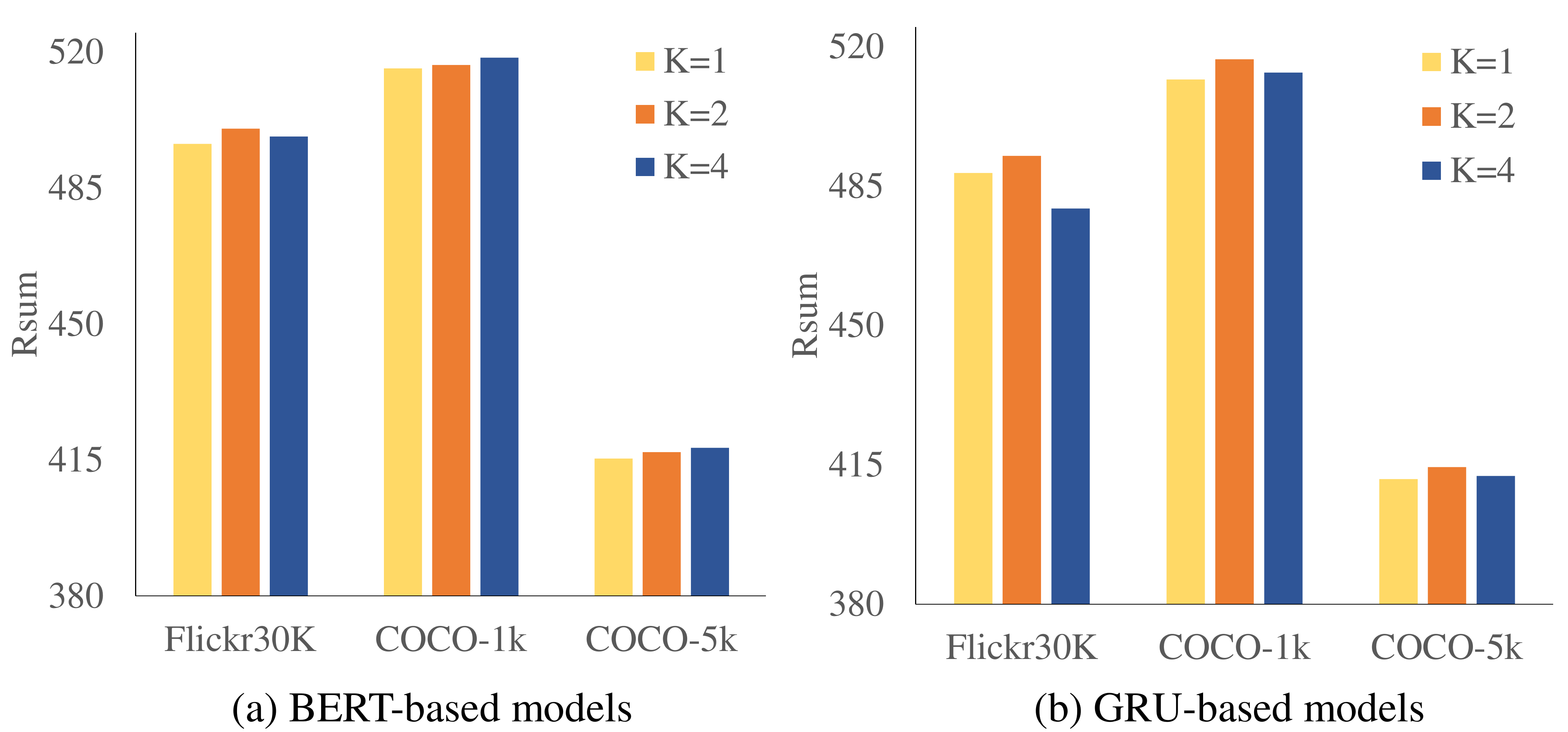}}
\caption{How Rsum goes as K grows for different test sets considering (a) models with BERT and (b) models with GRU.}
\label{fig6}

\end{figure}

\begin{figure}[t]
\centerline{\includegraphics[width=0.95\linewidth]{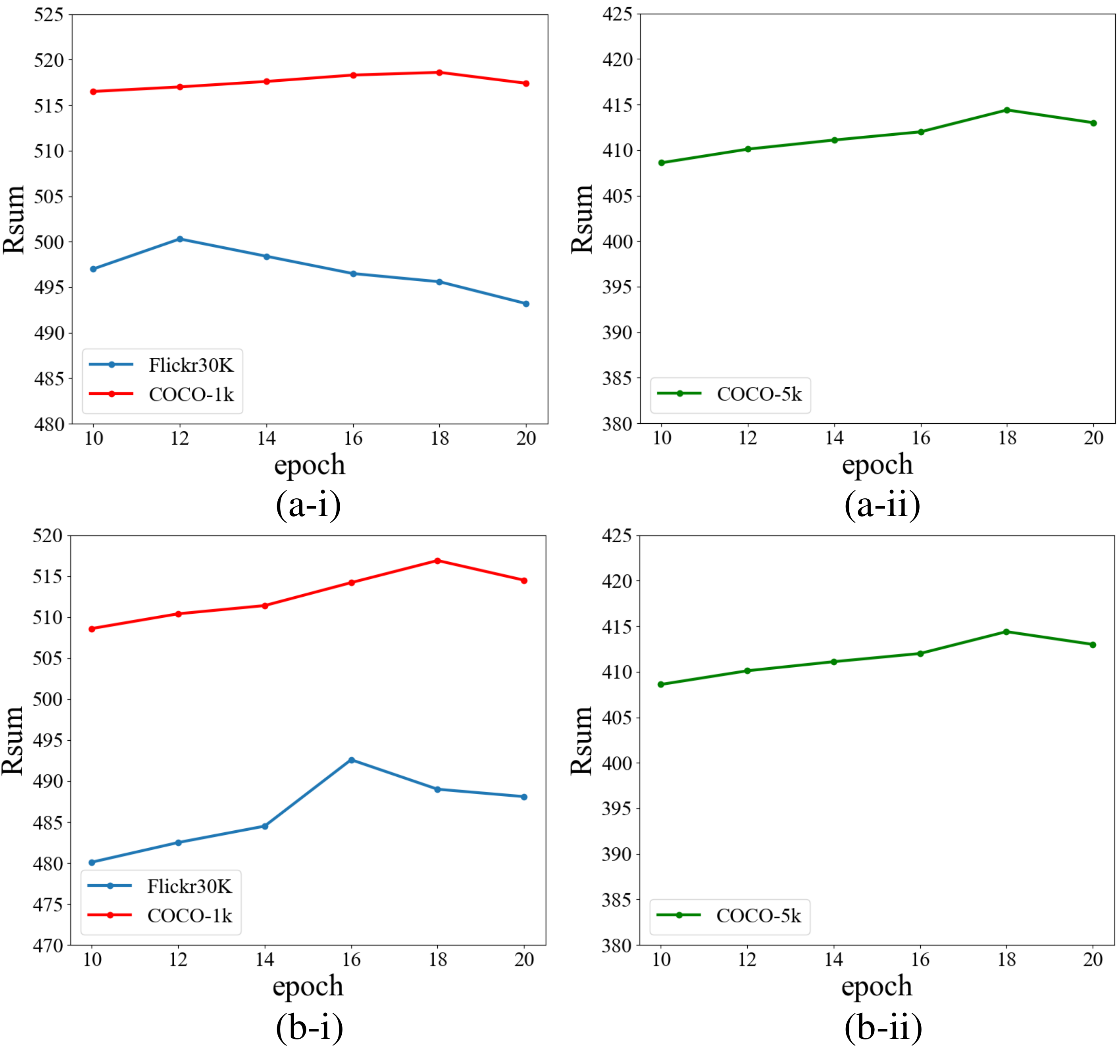}}
\caption{Rsum results for models with different training epochs on a) BERT-based models, b) GRU-based models, and i) Flickr30K and COCO-1k test set, ii) COCO-5k test set.}
\label{fig7}
\end{figure}

\begin{figure}[ht] 
\centerline{\includegraphics[width=1.0\linewidth]{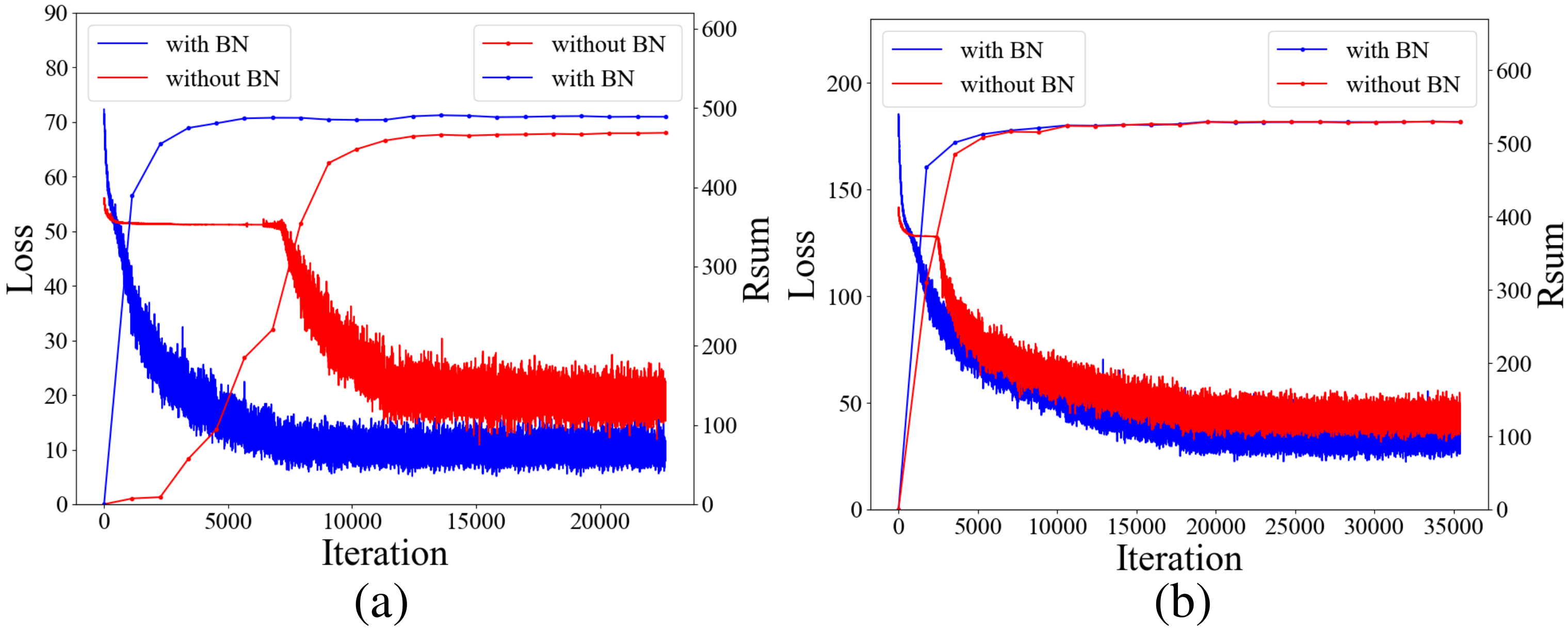}}
\caption{The plots of training losses and Rsum testing on a) Flickr30K validation set, b) COCO validation set.}
\label{fig8}
\end{figure}

\subsubsection{GATs in separate semantic relations module}

As seen in lines 1-4 and 9-12 in Table \ref{tab:sr-ablation}, a noticeable performance gain can be obtained with SSR for both single-path methods. Concretely, with SSR, $Rsum$ of the single global path increases from 396.4 to 409.0. The performance gain comes from learning global pixel-wise semantic relations. While $Rsum$ of the single regional path has much more significant growth of 52.5 from 439.1. This comes from the enhancement of regional relations learning realized by our graph attention module. 

In Section \ref{B}, we construct a global visual graph $G_F$ and regional visual graph $G_R$ and utilize the GATs to enhance global pixel-wise relations and regional object-wise relations. Specifically, for the image feature map directly extracted by CNNs, it has a grid-like structure, and the arrangement is relatively uniform. While GATs are better at processing graph-structured data, such as the image regions with irregular distribution as shown in Fig.~\ref{fig2}. Further, to illustrate the effectiveness of the GATs on capturing semantic relations, we visualize the attention using Flickr30K dataset. Considering that the final image representation should pay more attention to the salient objects in the raw images. Firstly we compute the similarities of the final image representation $\vec{I}$ and CNN-extracted global features $V_F$ or Faster-RCNN-extracted regional features $V_R$ with simply dot production. Thus, every area or every region has a similarity value with the final visual representation. Then we rank the values, thus the areas or regions with higher ranks are marked brighter in the attention maps, as shown in Fig.~\ref{fig5}. The left picture is the original image, the middle one is the image with salient regions highlighted, and the right one is the picture with some areas marked brighter. We choose the top-50 regions ranked by similarity score for regional attention visualization whicle for global attention, the raw image is cropped into $7\times7=49$ areas. Experiments are conducted whether using model 2 or model 4, as in Table \ref{tab:sr-ablation}. Four instances are given.

Focusing on the middle column, the regional attention pictures, the final image representation successfully concentrates on the images' salient objects. However, it is hard for global attention pictures to pay attention to the salient areas concerning the words in sentences. To sum up, as discussed in GAT\cite{gat}, what is more powerful about graph attention network is that it deals with the data of an irregular graph structure. Thus, when dealing with the regional visual graph, GATs help stress more salient nodes, which contribute more to the alignment with words in sentences. As for the global attention, although GATs can not pay attention to more significant areas, the relations-enhanced global features can still be used as a supplement to the full image information, and promote the later joint relations learning.

\subsubsection{GATs in joint semantic relations module}  

In Section \ref{C}, we learn the joint semantic relations by dint of GATs. With the strong ability of GATs in capturing node-level relations, the JSR successfully learns the region-global relations. As discussed before, for more profound attention in the graph attention module, we apply multi-head graph attention, and for different multi-head number $K$, the fusion process varies. As discussed in \cite{attention}, the use of multi-head attention enables each head to focus on different priorities, so more regional-global relations can be captured by the model. Fig.~\ref{fig6} records the Rsums with the change of $K$ on three different testing sets Flickr30K, COCO-1k, and COCO-5k using (a) BERT and (b) GRU. For Flickr30K, when $K$ equals 2, Rsum is the highest while for COCO-1k and COCO-5k, the top comes when $K$ is 4 for BERT and 2 for GRU. The difference between Flickr30K and COCO is that COCO has a richer training set, so even if the network gets deeper, the model can still learn more abundant relations. However, when applying GRU, the deficiency of masked pre-training limits the performance when $K$ gets larger.

\subsection{Analysis on Training Process}\label{V-C}

\subsubsection{Analysis of training epochs} 

In this part, to reduce the influence of underfitting and overfitting, we aim to find out the best training epochs for Flickr30K and MSCOCO, respectively. We conduct experiments using the full models whose configuration is the same as number 8 and 16 in Table.~\ref{tab:sr-ablation}. It should be noticed that for each experiment, the learning rate declines by ten times every half of the total epochs. As seen, for Flickr30K test set, the curve peaks when the epoch is set to 12 for BERT and 16 for GRU. Moreover, for MSCOCO, as seen in both 1k and 5k test sets, the training epoch should be set to 18.

\subsubsection{Analysis on batch normalization}

When building the graph attention module in Section \ref{B}, we add a batch normalization at the end of the module. As discussed in \cite{bn}, a BN layer can prevent the gradient explosion or dispersion, improve the robustness of the model to different super parameters (learning rate, initialization), and make most of the activation functions far away from its saturation region. All these properties of BN can help us to train the network quickly and robustly. Furthermore, the critical point is that the BN changes the optimization problem and makes the optimization space very smooth. When training our model, the batch normalization layer benefits the training process. As shown in Fig.~\ref{fig8}, we record the process of loss decline and the $Rsum$ on the validation set with or without a BN layer on BERT-based models. The use of batch normalization makes the optimization more smooth, and the loss drops faster. What's more, it promotes the training process to converge earlier and boosts the final evaluation performances.

\section{Conclusion}
In this paper, we focus on the visual semantic relations learning for enhanced image-text matching. Further, a dual semantic relations attention network (DSRAN) with different kinds of relations modules applied to capture both the object-level semantic relations and global-regional semantic relations. The learned dual-relations-enhanced visual representations can better match their textual counterparts whose words are inherently related in both object level and global-region level, thus promoting the matching procedure. Quantitative experiments show our model's successful target-oriented designs, and such a model outperforms previous methods on the image-text matching task on the two widely used datasets MSCOCO and Flickr30K. Further, we do ablation studies proving the effectiveness of the two main modules targeting dual semantic relations learning. In the future, we are looking forward to applying this kind of dual semantic relations learning to more cross-modal tasks such as image captioning and visual question answering.

\end{document}